\documentstyle[aclap,covingtn]{article}

\title{A Maximum Entropy Approach to Identifying Sentence Boundaries}

\author{Jeffrey C. Reynar and Adwait Ratnaparkhi\thanks{\ \ The authors
would like to acknowledge the support of ARPA grant N66001-94-C-6043,
ARO grant DAAH04-94-G-0426 and NSF grant SBR89-20230.}\\ Department of
Computer and Information Science\\ University of Pennsylvania\\
Philadelphia, Pennsylvania, USA\\
\{jcreynar, adwait\}@unagi.cis.upenn.edu}

\begin{document}
\bibliographystyle{fullname}
\maketitle
\vspace{-0.5in}
\begin{abstract}
We present a trainable model for identifying sentence boundaries in
raw text. Given a corpus annotated with sentence boundaries, our model
learns to classify each occurrence of {\em .}, {\em ?}, and {\em !}\
as either a valid or invalid sentence boundary.  The training
procedure requires no hand-crafted rules, lexica, part-of-speech tags,
or domain-specific information.  The model can therefore be trained
easily on any genre of English, and should be trainable on any other
Roman-alphabet language.  Performance is comparable to or better than
the performance of similar systems, but we emphasize the simplicity of
retraining for new domains.
\end{abstract}

\section{Introduction}

The task of identifying sentence boundaries in text has not received
as much attention as it deserves.  Many freely available natural
language processing tools require their input to be divided into
sentences, but make no mention of how to accomplish this
(e.g. \cite{brill:ai,collins}). Others perform the division implicitly
without discussing performance (e.g. \cite{cutting}).

On first glance, it may appear that using a short list of
sentence-final punctuation marks, such as {\em .}, {\em ?}, and {\em
!}, is sufficient. However, these punctuation marks are not used
exclusively to mark sentence breaks. For example, embedded quotations
may contain any of the sentence-ending punctuation marks and {\em .}\
is used as a decimal point, in e-mail addresses, to indicate ellipsis
and in abbreviations. Both {\em !}\ and {\em ?}\ are somewhat less
ambiguous but appear in proper names and may be used multiple times
for emphasis to mark a single sentence boundary.

Lexically-based rules could be written and exception lists used to
disambiguate the difficult cases described above. However, the lists
will never be exhaustive, and multiple rules may interact badly since
punctuation marks exhibit absorption properties. Sites which logically
should be marked with multiple punctuation marks will often only have
one (\cite{nunberg} as summarized in \cite{white}). For example, a
sentence-ending abbreviation will most likely not be followed by an
additional period if the abbreviation already contains one (e.g. note
that {\em D.C} is followed by only a single {\em .} in {\em The
president lives in Washington, D.C.}).

As a result, we believe that manually writing rules is not a good
approach. Instead, we present a solution based on a maximum entropy
model which requires a few hints about what information to use and a
corpus annotated with sentence boundaries. The model trains easily and
performs comparably to systems that require vastly more
information. Training on $39441$ sentences takes $18$ minutes
on a Sun Ultra Sparc and disambiguating the boundaries in a single
{\em Wall Street Journal} article requires only $1.4$ seconds.

\section{Previous Work}
To our knowledge, there have been few papers about identifying
sentence boundaries. The most recent work will be described in
\cite{ph:cl}. There is also a less detailed description of Palmer and
Hearst's system, SATZ, in \cite{ph:satz}.\footnote{We recommend these
articles for a more comprehensive review of sentence-boundary
identification work than we will be able to provide here.} The SATZ
architecture uses either a decision tree or a neural network to
disambiguate sentence boundaries. The neural network achieves 98.5\%
accuracy on a corpus of {\em Wall Street Journal} articles using a
lexicon which includes part-of-speech (POS) tag information. By
increasing the quantity of training data and decreasing the size of
their test corpus, Palmer and Hearst achieved performance of 98.9\%
with the neural network. They obtained similar results using the
decision tree. All the results we will present for our algorithms are
on their initial, larger test corpus.

In \cite{riley}, Riley describes a decision-tree based approach to the
problem. His performance on the Brown corpus is 99.8\%, using a model
learned from a corpus of 25 million words. Liberman and Church suggest
in \cite{lc:textanalysis} that a system could be quickly built to
divide newswire text into sentences with a nearly negligible error
rate, but do not actually build such a system.

\section{Our Approach}
We present two systems for identifying sentence boundaries. One is
targeted at high performance and uses some knowledge about the
structure of English financial newspaper text which may not be
applicable to text from other genres or in other languages. The other
system uses no domain-specific knowledge and is aimed at being
portable across English text genres and Roman alphabet languages.

Potential sentence boundaries are identified by scanning the text for
sequences of characters separated by whitespace (tokens) containing
one of the symbols {\em !}, {\em .}\ or {\em ?}. We use information
about the token containing the potential sentence boundary, as well as
contextual information about the tokens immediately to the left and to
the right. We also conducted tests using wider contexts, but
performance did not improve.

We call the token containing the symbol which marks a putative
sentence boundary the Candidate. The portion of the Candidate
preceding the potential sentence boundary is called the Prefix and the
portion following it is called the Suffix. The system that focused on
maximizing performance used the following hints, or contextual
``templates'':
\begin{itemize}
\item The Prefix
\item The Suffix
\item The presence of particular characters in the Prefix or Suffix
\item Whether the Candidate is an honorific (e.g. {\em Ms., Dr., Gen.}) 
\item Whether the Candidate is a corporate designator (e.g. {\em Corp., S.p.A.,
L.L.C.}) 
\item Features of the word left of the Candidate
\item Features of the word right of the Candidate
\end{itemize}

The templates specify only the form of the information. The {\em
exact} information used by the maximum entropy model for the potential
sentence boundary marked by {\em .}\ in {\em Corp.} in
Example~\ref{sample} would be: PreviousWordIsCapitalized, Prefix={\em
Corp}, Suffix=NULL, PrefixFeature=CorporateDesignator.

\begin{example}
\label{sample} ANLP Corp. chairman Dr. Smith resigned.
\end{example}

The highly portable system uses only the identity of the Candidate and
its neighboring words, and a list of abbreviations induced from the
training data.\footnote{A token in the training data is considered an
abbreviation if it is preceded and followed by whitespace, and it
contains a {\em .} that is {\em not} a sentence boundary.}
Specifically, the ``templates'' used are:
\begin{itemize}
\item The Prefix
\item The Suffix
\item Whether the Prefix or Suffix is on the list of induced abbreviations
\item The word left of the Candidate
\item The word right of the Candidate
\item Whether the word to the left or right of the Candidate is on the
list of induced abbreviations
\end{itemize}

The information this model would use for Example~\ref{sample}
would be: PreviousWord={\em ANLP}, FollowingWord={\em chairman},
Prefix={\em Corp}, Suffix=NULL, PrefixFeature=InducedAbbreviation.

The abbreviation list is automatically produced from the training
data, and the contextual questions are also automatically generated by
scanning the training data with question templates.  As a result, no
hand-crafted rules or lists are required by the highly portable system
and it can be easily re-trained for other languages or text genres.

\section{Maximum Entropy}
\label{maxentdescription}

The model used here for sentence-boundary detection is based on the
maximum entropy model used for POS tagging in \cite{ratnaparkhi}.  For
each potential sentence boundary token ({\em .}, {\em ?}, and {\em
!}), we estimate a joint probability distribution $p$ of the token and
its surrounding context, both of which are denoted by $c$, occurring
as an actual sentence boundary.  The distribution is given by:
\mbox{$p(b,c)=\pi\prod_{j=1}^k\alpha_j^{f_j(b, c)}$}, where $b \in \{
{\tt no}, {\tt yes} \}$, where the $\alpha_j$'s are the unknown
parameters of the model, and where each $\alpha_j$ corresponds to a
$f_j$, or a {\em feature}.  Thus the probability of seeing an actual
sentence boundary in the context $c$ is given by $p({\tt yes}, c)$.

The contextual information deemed useful for sentence-boundary
detection, which we described earlier, must be encoded using features.
For example, a useful feature might be:
\[ f_j(b, c) = 
\left\{ 
\begin{array}{ll}
1&\mbox{if Prefix($c$) = {\em Mr} \& $b$ = {\tt no}} \\
0&\mbox{otherwise}
\end{array}
\right. \] This feature will allow the model to discover that the
period at the end of the word {\em Mr.} seldom occurs as a sentence
boundary.  Therefore the parameter corresponding to this feature will
hopefully boost the probability $p({\tt no},c)$ if the Prefix is {\em
Mr}.  The parameters are chosen to maximize the likelihood of the
training data using the {\em Generalized Iterative Scaling}
\cite{scaling} algorithm.

The model also can be viewed under the Maximum Entropy framework, in
which we choose a distribution $p$ that maximizes the entropy $H(p)$
\[ H(p) = -\sum p(b,c) \log p(b,c) \]
under the following constraints:
\[ \sum p(b,c) f_j(b,c) = \sum \tilde{p}(b,c) f_j(b,c), 1 \leq j \leq k \]
where $\tilde{p}(b,c)$ is the observed distribution of 
sentence-boundaries and contexts in the training data.
As a result, the model in practice tends not to commit towards a particular
outcome ({\tt yes} or {\tt no}) unless it has seen sufficient evidence for
that outcome; it is maximally uncertain beyond meeting the evidence.

All experiments use a simple decision rule to classify each 
potential sentence boundary: a potential sentence boundary is an 
actual sentence boundary if and only if $p({\tt yes} | c) > .5$,
where 
\[ p({\tt yes} | c ) = { p({\tt yes}, c) \over 
{  p({\tt yes}, c) + p({\tt no}, c) } } \]
and where $c$ is the context including the potential sentence boundary.

\section{System Performance}

\begin{table}[ht]
\centering
\begin{tabular}{|l|r|r|} \hline
                   & WSJ    & Brown  \\ \hline
Sentences          & 20478  & 51672  \\ \hline
Candidate P. Marks & 32173  & 61282  \\ \hline
Accuracy           & 98.8\% & 97.9\% \\ \hline
False Positives    & 201    & 750    \\ \hline
False Negatives    & 171    & 506    \\ \hline
\end{tabular}
\caption{Our best performance on two corpora.}
\label{maxperf}
\end{table}

We trained our system on $39441$ sentences ($898737$ words) of {\em
Wall Street Journal} text from sections $00$ through $24$ of the
second release of the Penn Treebank\footnote{We did not train on files
which overlapped with Palmer and Hearst's test data, namely sections
03, 04, 05 and 06.}  \cite{treebank}. We corrected punctuation
mistakes and erroneous sentence boundaries in the training
data. Performance figures for our best performing system, which used a
hand-crafted list of honorifics and corporate designators, are shown
in Table~\ref{maxperf}. The first test set, WSJ, is Palmer and
Hearst's initial test data and the second is the entire Brown
corpus. We present the Brown corpus performance to show the importance
of training on the genre of text on which testing will be
performed. Table~\ref{maxperf} also shows the number of sentences in
each corpus, the number of candidate punctuation marks, the accuracy
over potential sentence boundaries, the number of false positives and
the number of false negatives. Performance on the WSJ corpus was, as
we expected, higher than performance on the Brown corpus since we
trained the model on financial newspaper text.

Possibly more significant than the system's performance is its
portability to new domains and languages. A trimmed down system which
used no information except that derived from the training corpus
performs nearly as well, and requires no resources other than a
training corpus. Its performance on the same two corpora is shown in
Table~\ref{portperf}.

\begin{table}[ht]
\centering
\begin{tabular}{|c|c|c|c|} \hline
Test    &            & False     & False      \\
Corpus	&   Accuracy & Positives & Negatives  \\ \hline
WSJ     &   98.0\%   & 396       & 245        \\ \hline
Brown   &   97.5\%   & 1260      & 265        \\ \hline
\end{tabular}
\caption{Performance on the same two corpora using the highly portable
system.}
\label{portperf}
\end{table}

\begin{table*}
\centering
\begin{tabular}{|c|c|c|c|c|c|c|c|} \hline
         & \multicolumn{7}{c|}{Number of sentences in training corpus}\\ \cline{2-8}
              & 500    & 1000   & 2000   & 4000   & 8000   & 16000  & 39441 \\ \hline
Best performing & 97.6\% & 98.4\% & 98.0\% & 98.4\% & 98.3\% & 98.3\% & 98.8\% \\ \hline
Highly portable & 96.5\% & 97.3\% & 97.3\% & 97.6\% & 97.6\% & 97.8\% & 98.0\% \\ \hline
\end{tabular}
\caption{Performance on {\em Wall Street Journal} test data as a
function of training set size for both systems.}
\label{trainingsize}
\end{table*}

Since  $39441$ training sentences is considerably more than
might exist in a new domain or a language other than English, we
experimented with the quantity of training data required to maintain
performance. Table~\ref{trainingsize} shows performance on the WSJ
corpus as a function of training set size using the best performing
system and the more portable system. As can seen from the table,
performance degrades as the quantity of training data decreases, but
even with only $500$ example sentences performance is better than the
baselines of $64.0\%$ if a sentence boundary is guessed at every
potential site and $78.4\%$ if only token-final instances of
sentence-ending punctuation are assumed to be boundaries.

\section{Conclusions}

We have described an approach to identifying sentence boundaries which
performs comparably to other state-of-the-art systems that require
vastly more resources.  For example, Riley's performance on the Brown
corpus is higher than ours, but his system is trained on the Brown
corpus and uses thirty times as much data as our system. Also, Palmer
\& Hearst's system requires POS tag information, which limits its use
to those genres or languages for which there are either POS tag lexica
or POS tag annotated corpora that could be used to train automatic
taggers.  In comparison, our system does not require POS tags or any
supporting resources beyond the sentence-boundary annotated corpus. It
is therefore easy and inexpensive to retrain this system for different
genres of text in English and text in other Roman-alphabet languages.
Furthermore, we showed that a small training corpus is sufficient for
good performance, and we estimate that annotating enough data to
achieve good performance would require only several hours of work, in
comparison to the many hours required to generate POS tag and lexical
probabilities.

\section{Acknowledgments}
We would like to thank David Palmer for giving us the test data he and
Marti Hearst used for their sentence detection experiments. We would
also like to thank the anonymous reviewers for their helpful insights.

\end{document}